\documentclass[10pt,twocolumn,letterpaper]{article}

\usepackage{3dv}
\usepackage{times}
\usepackage{epsfig}
\usepackage{graphicx}
\usepackage{amsmath}
\usepackage{amssymb}

\usepackage{color}

\threedvfinalcopy 

\def\threedvPaperID{188} 

\ifthreedvfinal\fi
\begin{document}

\title{Fast Single Shot Detection and Pose Estimation}

\author{Patrick Poirson$^1$, Phil Ammirato$^1$, Cheng-Yang Fu$^1$, Wei Liu$^1$, Jana Ko\v{s}eck\'{a}$^2$, Alexander C. Berg$^1$\\
$^1$UNC Chapel Hill $^2$George Mason University\\  
$^1$201 S. Columbia St., Chapel Hill, NC 27599 $^2$4400 University Dr, Fairfax, VA 22030\\
{\tt\small $^1$\{poirson,ammirato,cyfu,aberg\}@cs.unc.edu, $^2$kosecka@gmu.edu}
}


\maketitle

\begin{abstract}
   For applications in navigation and robotics, estimating the 3D pose of objects is as important as detection.  Many approaches to pose estimation rely on detecting or tracking parts or keypoints~\cite{pepik2012teaching,xiang2014beyond}.  In this paper we build on a recent state-of-the-art convolutional network for sliding-window detection~\cite{liu2016ssd} to provide detection and rough pose estimation in a single shot, without intermediate stages of detecting parts or initial bounding boxes. While not the first system to treat pose estimation as a categorization problem, this is the first attempt to combine detection and pose estimation at the same level using a deep learning approach.  The key to the architecture is a deep convolutional network where scores for the presence of an object category, the offset for its location, and the approximate pose are all estimated on a regular grid of locations in the image. The resulting system is as accurate as recent work on pose estimation (42.4\% 8 View mAVP on Pascal 3D+~\cite{xiang2014beyond} ) and significantly faster (46 frames per second (FPS) on a TITAN X GPU). This approach to detection and rough pose estimation is fast and accurate enough to be widely applied as a pre-processing step for tasks including high-accuracy pose estimation, object tracking and localization, and vSLAM.
\end{abstract}


\begin{figure}[t]
\centering
\includegraphics[width=0.3\textwidth]{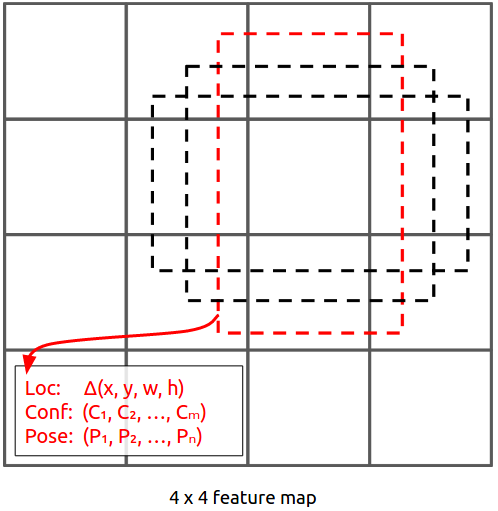}
\caption{\textbf{Default Box Predictions.} At each of a fixed set of locations, indicated by solid boxes, predictions are made for a collection of ``default" boxes of different aspect ratios.  In the SSD detector, for each default box a score for each object categery (Conf) is predicted, as is an offset in the positioning of the box (Loc).  This work adds a prediction for the pose of an object in the default box, represented by one of a fixed set of possible poses, $P_1\ldots P_n$.}

\label{fig:anchors}
\end{figure}

\section{Introduction}

Detecting and estimating the pose of objects in everyday scenes is a basic capability needed for automatic visual understanding of the world around us, including augmented reality applications, surveillance, navigation, manipulation, and robotics in general. For these applications it is necessary for systems to perceive the pose of objects in addition to the presence and location of an object. A wide range of recent work on object detection in computer vision offers one way to attack this problem. It is now possible, and relatively straightforward, to build detectors for some object {\em categories}, for example opened laptops. While classical approaches to estimating the pose of objects worked on {\em instances} of specific objects, often by finding keypoints whether through detection or tracking, object category detection offers another possible approach. Detecting categories of objects instead of tracking a specific object provides robustness, allows success with previously unseen objects, and can act as initialization or re-initialization for tracking-based approaches to pose estimation. Recently some approaches for object category detection have been extended to estimate the pose of objects, usually in a two stage approach, first detecting objects and then classifying the bounding box around a detected object into a set of coarse pose categories.

In this paper we push further along the direction of category-level object detection and pose estimation by integrating a pose estimation computation directly into a new, state-of-the-art, object detector~\cite{liu2016ssd}. This detector uses a deep convolutional network to predict both the presence of a category and adjustments to a bounding box at each of a large set of possible bounding box locations in an image (depicted as the dashed boxes in Fig.~\ref{fig:anchors}). We extend these predictions to include object pose (the red dashed box in Fig.~\ref{fig:anchors}). To our knowledge this is the first time a deep-network based approach has been used to perform object detection and pose estimation at the same level. Previous work combining the two built on the deformable parts (DPM) approach~\cite{xiang2014beyond,pepik2012teaching} adding pose estimation as part of a DPM-based structured prediction model.

By integrating pose estimation directly into a very fast object detection pipeline, instead of adding it as a secondary classification stage, the resulting detection and pose estimation system is very fast, up to 46 frames per second (FPS) on a Titan X GPU, at least 7x faster than Faster R-CNN~\cite{ren2015faster} and close to 100 times faster than R-CNN~\cite{rcnn}. Those subroutines are used as pre-processing in almost all other recent category detection and pose estimation pipelines, so the entire compute time is even longer. This speed allows the new system to be run as a pre-processing step for other computations, e.g. initializing a more computationally expensive high-accuracy pose estimation system based on dense correspondence, or adding robustness to a real-time object tracking system, or as part of a visual simultaneous localization and mapping (visual SLAM) tool. Employing coarse pose estimation as a pre-processing step makes sense, and the relatively low compute requirements of our approach make this much more tenable than previous work.

We present the new detection and pose estimation pipeline in Sec.~\ref{sec:model}, and evaluate it on the standard Pascal 3D+ dataset~\cite{xiang2014beyond} in Sec.~\ref{sec:p3d}. The Pascal 3D+ data comes from the Pascal VOC detection dataset, which was collected from web images. In order to further validate our approach, we also evaluate on a new collection of images of everyday environments taken with a low-quality RGB camera, Sec.~\ref{sec:philDataset}. There we compare models trained on Pascal 3D+, models trained on Pascal 3D+ and fine tuned on the new data, and models trained only on the new dataset.


\section{Previous Work}

Pose estimation is a well-studied problem in computer vision. Initially, research on the topic focused primarily on instance-level pose estimation~\cite{instance1,instance2,instance3,instance4}. In instance-level pose estimation, the model is trained on each specific object instance. Given enough training images of a single instance, the problem essentially reduces to constructing a 3D model of the instance and then finding the pose of the 3D model that best matches the test instance. 

More recently, however, the field has shifted to working on category-level pose estimation~\cite{savarese20073d,glasner2011viewpoint, ghodrati20142d}. This presents many new challenges; most notably, the model must learn how the pose changes with the appearance of the object while simultaneously handling the intra-category variation of instances within a category.

One of the earlier attempts at category-level detection and pose estimation was done by Pepik~\etal~\cite{pepik2012teaching}; they formulate the joint object detection and pose estimation problem as a structured prediction problem. They propose a structural SVM, which predicts both the bounding box and pose jointly. This will be referred to as DPM-VOC+VP. Similarly, Xiang~\etal~\cite{xiang2014beyond} extend the original DPM method such that each mixture component represents a different azimuth section. This will be referred to as VDPM.

\begin{figure*}[t]
\centering
\includegraphics[width=0.7\textwidth]{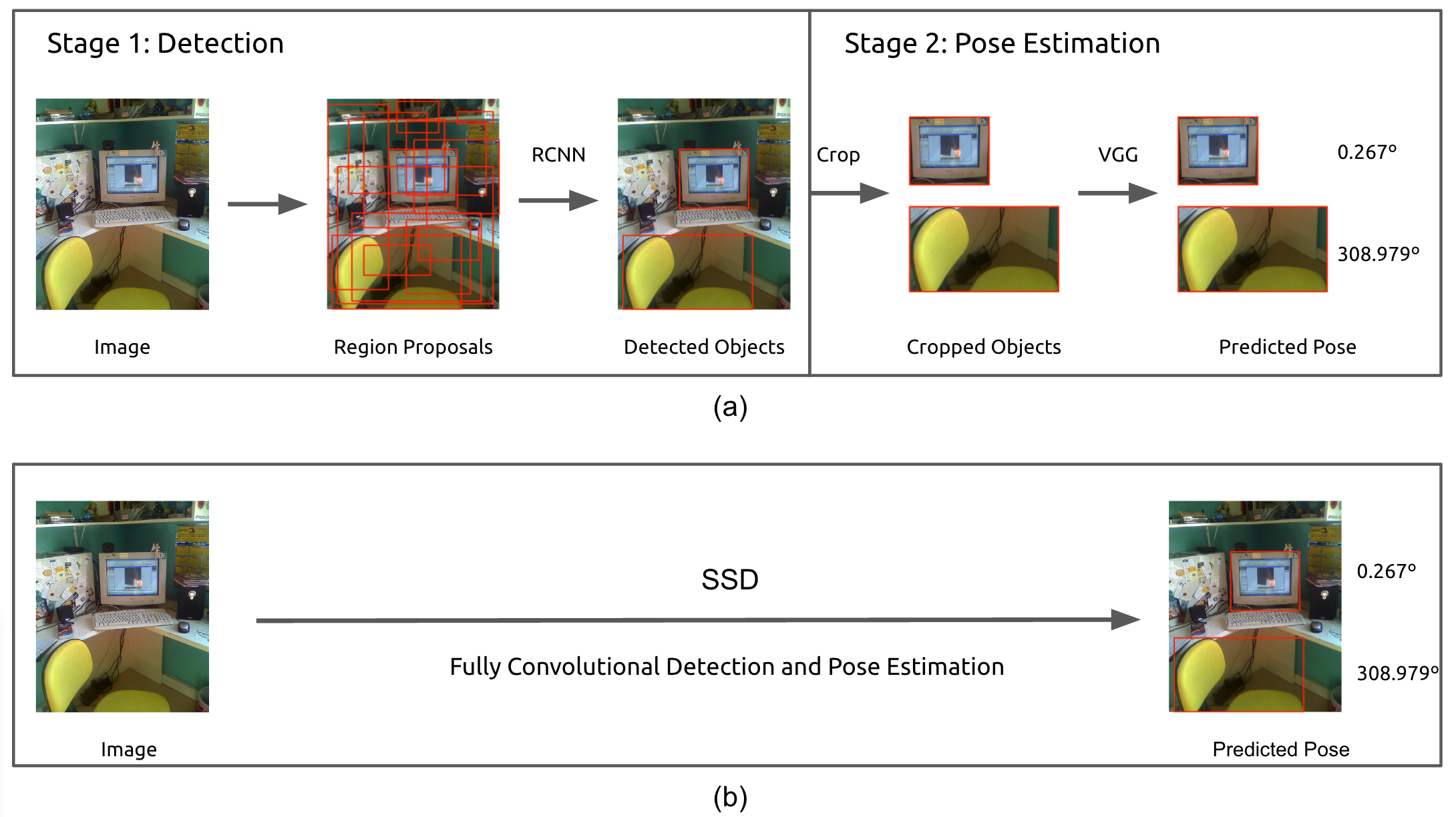}
\caption{\textbf{Two-stage vs. Proposed.} (a) The two-stage approach separates the detection and pose estimation steps. After object detection, the detected objects are cropped and then processed by a separate network for pose estimation. This requires resampling the image at least three times: once for region proposals, once for detection, and once for pose estimation. (b) The proposed method, in contrast, requires no resampling of the image and instead relies on convolutions for detecting the object and its pose in a single forward pass. This offers a large speed up because the image is not resampled, and computation for detection and pose estimation is shared.}

\label{fig:twoStage}
\end{figure*}

While these earlier methods sought to perform joint detection and pose estimation, many of the more recent deep learning approaches to pose estimation have separated the problem into two stages (Fig. \ref{fig:twoStage} (a)). Stage one requires using an off the shelf detector:~\cite{Su_2015_ICCV} and~\cite{glasner2011viewpoint} use RCNN~\cite{rcnn}, but more recent detectors such as Faster RCNN~\cite{ren2015faster} could naturally be substituted (although the change in object proposal might affect the accuracy). Stage two requires cropping the image for the detected regions, which are individually processed by a separate network (e.g., Alexnet~\cite{krizhevsky2012imagenet} or VGG~\cite{simonyan2014very}) fine-tuned for pose estimation. 

Tulisani~\etal~\cite{tulsiani2015viewpoints} treat pose estimation as a classification problem and use a Convolutional Neural Network (CNN) for predicting the pose. Their pose estimation network is initialized with VGG~\cite{simonyan2014very} and finetuned for pose estimation using ground truth annotations from Pascal 3D+. In order to perform object detection and pose estimation they first use R-CNN~\cite{rcnn} to detect objects and then use the detected regions as input to their pose estimation network. Their results are the current state of the art for detection and pose estimation on Pascal 3D+. In Sec.~\ref{sec:experiments}, we compare our detection and pose estimation model to theirs and demonstrate similar accuracy. Additionally, in table~\ref{tbl:speed} we show a timing comparison between our proposed model and theirs. This table shows that even if their method used the fastest region based detector (Faster RCNN) our method still provides a 46x increase in speed with similar accuracy.

Similarly, Su~\etal~\cite{Su_2015_ICCV} use the same formulation as~\cite{tulsiani2015viewpoints} for pose estimation. However, with their rendering pipeline they are able to generate more than 2 million synthetic images with ground truth pose annotations. They use these synthetic images to train a pose estimation network, which surpasses~\cite{tulsiani2015viewpoints} in viewpoint estimation accuracy. Finally, they too perform two-stage detection and pose estimation with R-CNN detections. In Sec.~\ref{sec:experiments}, we compare our detection and pose estimation model to~\cite{Su_2015_ICCV} and show significant increases in accuracy.  

Even if a two-staged pipeline for detection and pose estimation uses a fast detector (e.g., Faster RCNN~\cite{ren2015faster}) in stage one, it will still be slower than performing joint detection and pose estimation, since the detected objects must be cropped and then processed by a separate network. In contrast, the method presented in this paper is a joint detector and pose estimator, which requires just one network and a single evaluation of the image (Fig.~\ref{fig:twoStage} (b)).

\section{Model}\label{sec:model}

For an input RGB image, a single evaluation of the model network is performed and produces scores for category, bounding box offset directions, and pose, for a constant number of boxes. These are filtered by non-max suppression to produce the final output. The network is a variant of the single shot detection (SSD)
network from~\cite{liu2016ssd} with additional outputs for pose. Here we present the network's design choices, structure of the outputs, and training.

\begin{figure*}
\centering
\includegraphics[width=\textwidth]{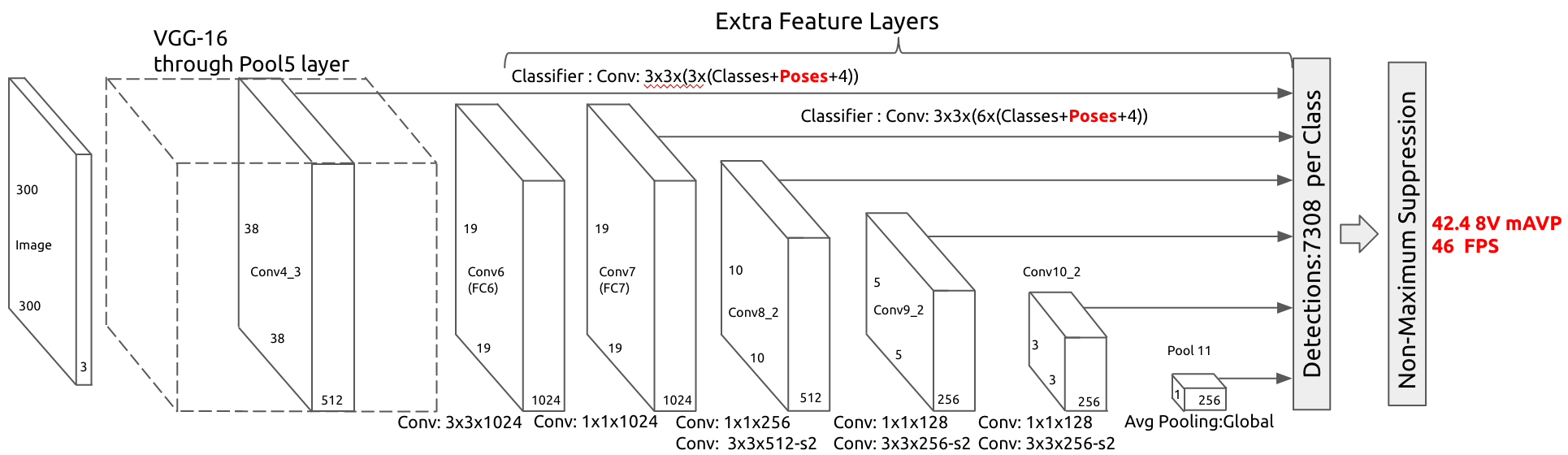}
\caption{\textbf{Our Model Architecture.} Detection and pose estimation network for the ``Share 300'' model that shares a single pose prediction across all categories at each location and resizes images to 300x300 before using them as input to the network.  Feature maps are added in place of the final layers of a VGG-16 network and small convolutional filters produce estimates for class, pose, and bounding box offsets that are processed through non-max suppression to make the final detections and pose estimates. Red indicates additions to the architecture of SSD~\cite{liu2016ssd}}
\label{fig:architecture}
\end{figure*}

An SSD-style detector~\cite{liu2016ssd} works by adding a sequence of feature maps of progressively decreasing spatial resolution to an image classification network such as VGG~\cite{simonyan2014very}. These feature layers replace the last few layers of the image classification network, and 3x3 and 1x1 convolutional filters are used to transform one feature map to the next along with max-pooling. See Fig.~\ref{fig:architecture} for a depiction of the model.

Predictions for a regularly spaced set of possible detections are computed by
applying a collection of 3x3 filters to channels in one of the feature layers.
Each 3x3 filter produces one value at each location, where the outputs are either classification scores, localization offsets, and, in our case, discretized pose predictions for the object (if any) in a box. See Fig. \ref{fig:anchors}. Note that different sized detections are produced by different feature layers instead of taking the more traditional approach of resizing the input image or predicting different sized detections from a single feature layer.

We take one of two different approaches for pose predictions, either sharing outputs for pose across all the object categories ({\em share}) or having separate pose outputs for each object category ({\em separate}).  One output is added for each of $N_\theta$ possible poses. With $N_c$ categories of objects, there are $N_c \times N_\theta$ pose outputs for the separate model and $N_\theta$ pose outputs for the share model. While we do add a 3x3 filter for each of the pose outputs, this added cost is relatively small and the original SSD pipeline is quite fast, so the result is still faster than two stage approaches that rely on a (often slower) detector followed by a separate pose classification stage. See Fig.~\ref{fig:twoStage} (a).

\subsection{Pose Estimation Formulation}
There are a number of design choices for a joint detection and pose estimation method. This section details three particular design choices, and Sec.~\ref{sec:p3d_tr_choice} shows justifications for them through experimental results.

One important choice is in how the pose estimation task is formulated. A possibility is to train for continuous pose estimation and formulate the problem as a regression. However, in this work we discretize the pose space into $N_\theta$ disjoint bins and formulate the task as a classification problem. Doing so not only makes the task feasible (since both the quantity and consistency of pose labels is not high enough for continuous pose estimation), but also allows us to measure the confidence of our pose prediction. Furthermore, discrete pose estimation still presents a very challenging problem.

Another design choice is whether to predict poses separately for the $N_c$ object classes or to use the same weights to predict poses for all classes. Sec.~\ref{sec:p3d_tr_choice} assess these options.

The final design choice is the resolution of the input image. Specifically, we consider two resolutions for input: 300x300 and 500x500. In Sec.~\ref{sec:p3d_tr_choice}, we compare models trained on both resolutions. Using 500x500 images should provide better accuracy because higher resolution images provide finer details that assist in determining the pose. However, processing 500x500 images reduces the speed of computation. Previous deep learning pose estimation formulations~\cite{tulsiani2015viewpoints,Su_2015_ICCV} crop the detected regions then resize the region to a fixed size of 227x227. This has the advantage of always providing a higher res view of the object. However, with our 300x300 model, an object can take up far less than a 100x100 region in the image since there is no cropping step in our pipeline; therefore, we are fighting against a low resolution. As a result, the higher resolution helps combat this issue.

\subsection{Training}

The training objective is composed of a weighted sum of three losses: $L_{cls}$, $L_{loc}$, and $L_{pose}$ (the class, localization, and pose loss, respectively). In order to compute the loss and backpropagate through our network, we need to first match the ground truth detection and pose annotations to the appropriate default boxes. We follow the approach presented in~\cite{liu2016ssd} and match a default box with a ground box if their $IoU > 0.5$. As a result, one ground truth bounding box can be matched to multiple default boxes.

Let $N$ be the total number of default boxes matched to a ground truth box. Normalizing our training objective by $N$, our total loss is \begin{equation} L_{tot} = \frac{1}{N}(L_{cls} + \alpha_1L_{loc} + \alpha_2L_{pose}), \end{equation} where $\alpha_1$ is set to 1 and $\alpha_2$ is set to 1.5 through cross validation. Our $L_{cls}$ and $L_{pose}$ losses are both softmax losses, whereas our $L_{loc}$ loss is a Smooth L1 regression loss. Additionally, we adopt the same hard negative mining strategy as~\cite{liu2016ssd}. However, we do not include their full sampler. They sample patches from the image with a minimum overlap of [0.3, 0.5, 0.7, and 0.9] with a ground truth box. In contrast, we sample patches with overlaps of [0.7 and 0.9] because it's too difficult to predict the pose otherwise.

\section{Experiments}\label{sec:experiments}
To evaluate our model, we ran several experiments for detection and pose estimation on two different datasets: Pascal 3D+ and a dataset that we collected for detection and pose estimation in real-world scenes. We found that our SSD model performs comparably to the state-of-the-art two-stage pipeline methods; this is surprising, considering the 46x speedup that our method has.
We evaluate our results using the AVP metric proposed by Xiang~\etal~\cite{xiang2014beyond}. AVP is an extension of the standard AP metric used to evaluate object detection. For object detection, the AP metric labels a prediction as a true positive if its bounding box has $IoU > 0.5$ with the ground truth bounding box and the correct class label. AVP adds an additional requirement that the predicted pose label must also be correct. AVP is evaluated at different levels of discretization of the pose space into $N_\theta$ bins: 4, 8, 16, and 24 bins.

\subsection{Pascal 3D+ Dataset}\label{sec:p3d}
The Pascal 3D+ dataset~\cite{xiang2014beyond} is made up of images from the Pascal~\cite{everingham2015pascal} and ImageNet~\cite{russakovsky2015imagenet} datasets that have been labeled with both detection and continuous pose annotations for the 12 rigid object categories that appear in Pascal VOC12~\cite{everingham2015pascal} Train and Val set. In our experiments on this dataset, we follow the standard split and use the labeled Pascal Val images for evaluating our models.

The first set of experiments discussed below justify several design decisions selected for our base model. The second set of experiments compares our best models with previously published methods.

\subsubsection{Training Choices}\label{sec:p3d_tr_choice}
\textit{Share vs. Separate}

First, we compare training on shared and separate pose estimation (Table \ref{tbl:seperate}). The shared model, as anticipated, provides significant speed improvements over the separate model, as explained in Sec.~\ref{sec:model}. Surprisingly, however, we observe that our share model performs better than the separate method. We hypothesize that this is because training a separate pose estimator requires more data, and Pascal 3D+ only has an average of 3091 images per class. Based on these findings, the remaining experiments use only the share model.

\begin{table}[!htbp]
\begin{center}
\begin{tabular}{|l|c|c|c|c|}
\hline
Method & 4 View & 8 View & 16 View & 24 View \\
\hline\hline

\hline
Share 300 & 48.1 & 42.3 & 31.9 & 27.7 \\ 
\hline 
Separate 300 & 47.6 & 40.6 & 29.8 & 25.5 \\ 
\hline 
\end{tabular}
\end{center}
\caption{Share vs Seperate.}
\label{tbl:seperate}
\end{table}

\vspace{12pt}
\noindent\textit{300x300 vs. 500x500}

Our next experiment analyzes the effects of using higher resolution images for training. Table \ref{tbl:p3d} shows the performance of our models using 300x300 and 500x500 images on each object class. In all cases we achieve greater than 2\% improvement with the higher resolution 500x500 model for the average case. On the other hand, the 500x500 reduces the FPS from 46 to 17. Since our goal is to build a joint detector and pose estimator suitable for real-time systems, we use the 300x300 model in subsequent experiments.

%

\vspace{12pt}
\noindent\textit{Pascal + Imagenet vs. Pascal}

Recall that the Pascal 3D+ dataset provides detection and pose ground truths for images from Pascal and Imagenet. The test set consists only of Pascal images, which presents a problem, as there are more labeled images in the Imagenet dataset than the Pascal dataset. Nevertheless, pose labels are an expensive resource to collect, so we cannot afford to ignore the additionally labeled Imagenet images. Furthermore, our deep learning approach benefits from the additional data. Consequently, it is essential that we use the labels from both the Pascal and Imagenet images. 

If one naively trains on the union of the labeled Pascal and Imagenet images, then the results on the Pascal test set will be diminished by the presence of Imagenet images. To resolve this issue and counteract the effects of the Imagenet images, we replicate the Pascal images so that the number of Pascal and Imagenet images in our training set are equal, ensuring that the number of images from each are approximately the same in each sampled minibatch.

To verify the benefit of training with both the Pascal and Imagenet images, we trained a shared 300 model using both Pascal and Imagenet training images as well as a Shared model using only Pascal training images (Table~\ref{tbl:imnet}). As hypothesized, the absence of Imagenet images negatively affects the performance of our model; therefore, the remaining experiments are trained with both Pascal and Imagenet images. 

\begin{table}
\begin{center}
\begin{tabular}{|l|c|c|c|c|}
\hline
Method & 4 View & 8 View & 16 View & 24 View \\
\hline\hline

\hline
Share 300 & 48.1 & 42.3 & 31.9 & 27.7 \\ 
\hline 
Share 300 Pascal & 43.2 & 36.3 & 22.8 & 20.7 \\ 
\hline 
\end{tabular}
\end{center}
\caption{Training with vs. without ImageNet annotations.}
\label{tbl:imnet}
\end{table}

\vspace{12pt}
\noindent\textit{Fine-grained training vs. Coarse-grained training}

We also present results training on 24 bin pose discretization and testing on 4 and 8 bins (Table~\ref{tbl:24v}). Theoretically, the 24 bin model tested on a coarser discretization of angles would provide results equal to models trained directly for 4 or 8 bins. 

In our case we find that the 24 bin model performs comparably to the 4 and 8 bin models; however, in all cases the coarse-grained training performs better.

\begin{table}[!htbp]
\begin{center}
\begin{tabular}{|l|c|c|}
\hline
Method & 4 View & 8 View  \\
\hline\hline
Share 300 & 45.5 & 40.9 \\ 
\hline 
Share 300 24-V & 43.4 & 34.7 \\
\hline
Share 500 & 48.0 & 43.3  \\
\hline
Share 500 24-V & 47.0 & 38.0 \\
\hline
\end{tabular}
\end{center}
\caption{24 view model tested on other binnings.}
\label{tbl:24v}
\end{table}

\begin{table*}
\begin{center}
\begin{tabular}{| l || c | c | c | c | c | c | c | c | c | c | c || c |} 
\hline
Methods & aero & bicycle & boat & bus & car & chair & table & mbike & sofa & train & monitor & Avg.\\
\hline\hline

\multicolumn{13}{| c |}{Joint Object Detection and Pose Estimation (4 View AVP)}\\
\hline
VDPM~\cite{xiang2014beyond} & 34.6 & 41.7 & 1.5 & 26.1 & 20.2 & 6.8 & 3.1 & 30.4 & 5.1 & 10.7 & 34.7 & 19.5 \\
\hline
DPM-VOC+VP~\cite{pepik2012teaching} & 37.4 & 43.9 & 0.3 & 48.6 & 36.9 & 6.1 & 2.1 & 31.8 & 11.8 & 11.1 & 32.2 & 23.8 \\
\hline
RCNN+Alex~\cite{Su_2015_ICCV} & 54.0 & 50.5 & 15.1 & 57.1 & 41.8 & 15.7 & 18.6 & 50.8 & 28.4 & 46.1 & 58.2 & 39.7 \\
\hline
VpKps~\cite{tulsiani2015viewpoints} & 63.1 & 59.4 & 23.0 & 69.8 & \textbf{55.2} & \textbf{25.1} & 24.3 & 61.1 & \textbf{43.8} & 59.4 & 55.4 & 49.1 \\
\hline
Ours Share 300 & 63.6 & 54.7 & 25.0 & 67.7 & 47.3 & 10.8 & 38.5 & 59.4 & 41.8 & 65.0 & 55.8 & 48.1 \\ 
\hline 
Ours Share 500 & \textbf{64.6} & \textbf{62.1} & \textbf{26.8} & \textbf{70.0} & 51.4 & 11.3 & \textbf{40.7} & \textbf{62.7} & 40.6 & \textbf{65.9} & \textbf{61.2} & \textbf{50.7} \\ 
\hline\hline

\multicolumn{13}{| c |}{Joint Object Detection and Pose Estimation (8 View AVP)}\\
\hline

VDPM~\cite{xiang2014beyond} & 23.4 & 36.5 & 1.0 & 35.5 & 23.5 & 5.8 & 3.6 & 25.1 & 12.5 & 10.9 & 27.4 & 18.7 \\
\hline
DPM-VOC+VP~\cite{pepik2012teaching} & 28.6 & 40.3 & 0.2 & 38.0 & 36.6 & 9.4 & 2.6 & 32.0 & 11.0 & 9.8 & 28.6 & 21.5 \\
\hline
RCNN+Alex~\cite{Su_2015_ICCV} & 44.5 & 41.1 & 10.1 & 48.0 & 36.6 & 13.7 & 15.1 & 39.9 & 26.8 & 39.1 & 46.5 & 32.9 \\
\hline
VpKps~\cite{tulsiani2015viewpoints} & 57.5 & 54.8 & 18.9 & 59.4 & \textbf{51.5} & \textbf{24.7} & 20.5 & \textbf{59.5} & \textbf{43.7} & 53.3 & 45.6 & 44.5 \\
\hline
Ours Share 300 & 57.6 & 50.8 & \textbf{20.9} & 58.4 & 43.1 & 9.1 & 34.2 & 52.3 & 37.2 & 55.6 & 46.7 & 42.4 \\ 
\hline 
Ours Share 500 & \textbf{58.6} & \textbf{56.4} & 19.9 & \textbf{62.4} & 45.2 & 10.6 & \textbf{34.7} & 58.6 & 38.8 & \textbf{61.2} & \textbf{49.7} & \textbf{45.1} \\ 
\hline\hline

\multicolumn{13}{| c |}{Joint Object Detection and Pose Estimation (16 View AVP)}\\
\hline
VDPM~\cite{xiang2014beyond} & 15.4 & 18.4 & 0.5 & 46.9 & 18.1 & 6.0 & 2.2 & 16.1 & 10.0 & 22.1 & 16.3 & 15.6 \\
\hline
DPM-VOC+VP~\cite{pepik2012teaching} & 15.9 & 22.9 & 0.3 & 49.0 & 29.6 & 6.1 & 2.3 & 16.7 & 7.1 & 20.2 & 19.9 & 17.3 \\
\hline
RCNN+Alex~\cite{Su_2015_ICCV} & 27.5 & 25.8 & 6.5 & 45.8 & 29.7 & 8.5 & 12.0 & 31.4 & 17.7 & 29.7 & 31.4 & 24.2 \\
\hline
VpKps~\cite{tulsiani2015viewpoints} & \textbf{46.6} & \textbf{42.0} & 12.7 & \textbf{64.6} & \textbf{42.7} & \textbf{20.8} & 18.5 & 38.8 & \textbf{33.5} & \textbf{42.5} & 32.9 & \textbf{36.0} \\
\hline
Ours Share 300 & 45.4 & 33.4 & 13.7 & 52.9 & 32.9 & 5.3 & \textbf{27.2} & 38.8 & 27.3 & 37.4 & \textbf{36.2} & 31.9 \\ 
\hline 
Ours Share 500 & 45.9 & 39.6 & \textbf{14.0} & 54.0 & 35.4 & 7.4 & 26.4 & \textbf{40.4} & 29.2 & 41.5 & 35.8 & 33.6 \\ 
\hline\hline

\multicolumn{13}{| c |}{Joint Object Detection and Pose Estimation (24 View AVP)}\\
\hline
VDPM~\cite{xiang2014beyond} & 8.0 & 14.3 & 0.3 & 39.2 & 13.7 & 4.4 & 3.6 & 10.1 & 8.2 & 20.0 & 11.2 & 12.1 \\
\hline
DPM-VOC+VP~\cite{pepik2012teaching} & 9.7 & 16.7 & 2.2 & 42.1 & 24.6 & 4.2 & 2.1 & 10.5 & 4.1 & 20.7 & 12.9 & 13.6 \\
\hline
RCNN+Alex~\cite{Su_2015_ICCV} & 21.5 & 22.0 & 4.1 & 38.6 & 25.5 & 7.4 & 11.0 & 24.4 & 15.0 & 28.0 & 19.8 & 19.8 \\
\hline
VpKps~\cite{tulsiani2015viewpoints} & \textbf{37.0} & \textbf{33.4} & 10.0 & 54.1 & \textbf{40.0} & \textbf{17.5} & 19.9 & \textbf{34.3} & \textbf{28.9} & 43.9 & 22.7 & \textbf{31.1} \\
\hline
Ours Share 300 & 35.7 & 23.6 & \textbf{10.8} & 51.7 & 33.8 & 6.2 & \textbf{23.6} & 26.9 & 20.4 & \textbf{46.9} & \textbf{25.3} & 27.7 \\ 
\hline 
Ours Share 500 & 33.4 & 29.4 & 9.2 & \textbf{54.7} & 35.7 & 5.5 & 22.9 & 30.3 & 27.5 & 44.1 & 24.3 & 28.8 \\ 
\hline
\end{tabular}
\end{center}
\caption{Category specific results on Pascal 3D+.}
\label{tbl:p3d}
\end{table*}

\subsubsection{Comparison to state-of-the-art}\label{sec:sotaP3d}
Su~\etal~\cite{Su_2015_ICCV} represents the state-of-the-art on pose estimation while Tulisani~\etal~\cite{tulsiani2015viewpoints} represent the state-of-the-art for object detection and pose estimation on Pascal 3D+. Both methods follow a similar two-stage pipeline for object detection and pose estimation: the first stage uses RCNN for detection, and the second stage uses a CNN finetuned for pose estimation. 

In table~\ref{tbl:speed} we compare the speed of our model to ~\cite{tulsiani2015viewpoints}. We achieve a very significant 46x increase in speed, which opens up a wide range of settings in which the system can be used. Furthermore, in Table~\ref{tbl:p3d} we demonstrate comparable results in accuracy to~\cite{tulsiani2015viewpoints} and even surpass~\cite{Su_2015_ICCV}.   

It is important to note that~\cite{Su_2015_ICCV} renders more than 2 million synthetic images, which is approximately 50 times larger than the original Pascal 3D+ dataset. These additional images are particularly useful for deep learning methods. Therefore, we suspect that using a large of amount of similar synthesized data could improve the accuracy of the method presented here. We leave this as a potential future work.

\begin{figure*}[t]
\centering
\includegraphics[width=0.95\textwidth]{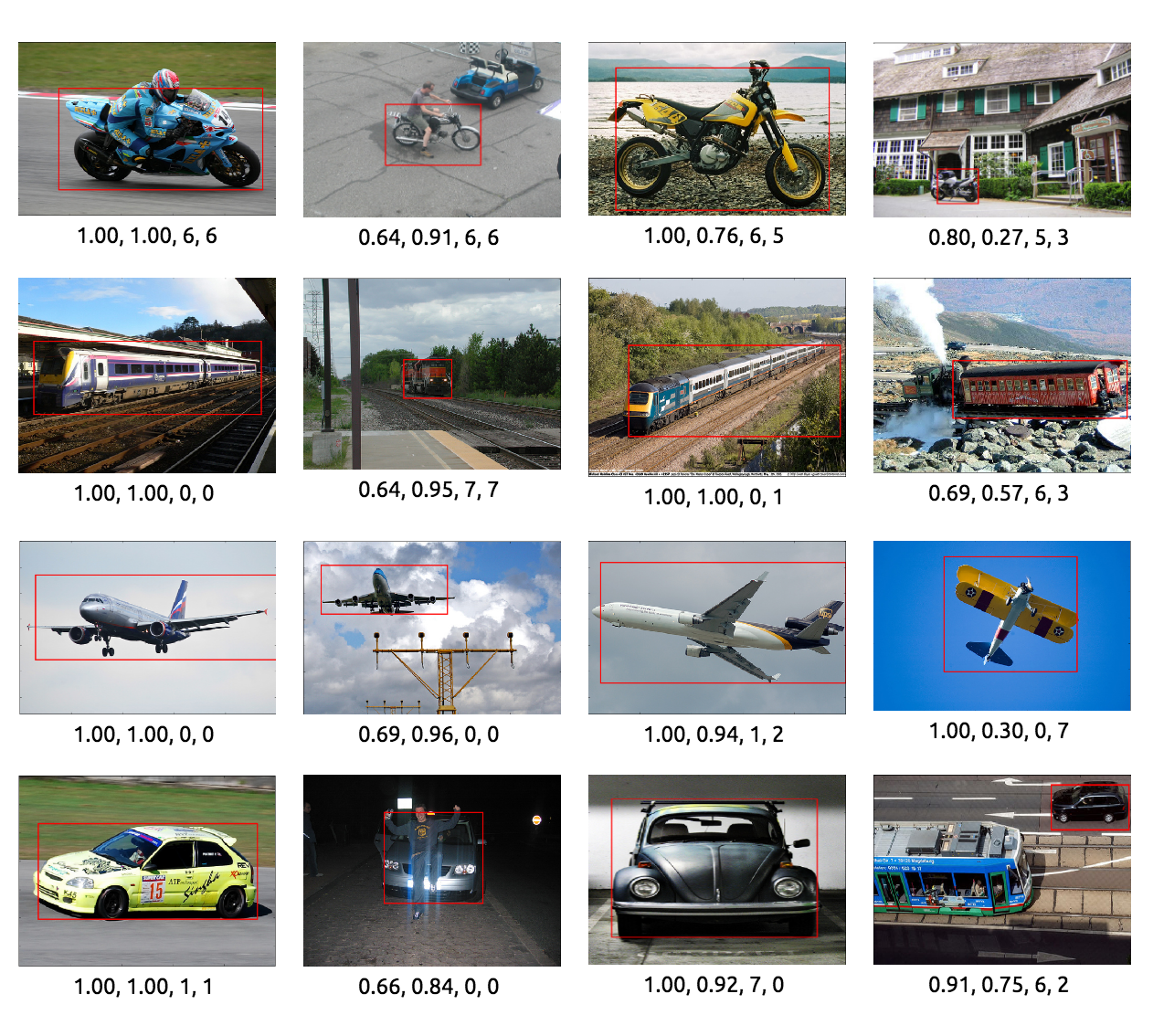}
\caption{\textbf{Pascal 3D+ Qualitative Results.} Results on 8 bin detection and pose estimation on the Pascal 3D+ dataset. Each image has a corresponding detection class confidence, pose confidence, predicted pose label, and ground truth pose label, respectively. Columns one and two show correct pose predictions with high and low detection scores; column three shows pose predictions that are off by one bin; and column four shows some difficult examples where our system fails. }

\label{fig:collage}
\end{figure*}

We also provide qualitative results on test images from Pascal 3D+ in Fig.~\ref{fig:collage}.

\begin{table}[!htbp]
\begin{center}
\begin{tabular}{|l|c|c|c|}
\hline
Method & Detector (FPS) & Pose (FPS) & Total (FPS)  \\
\hline\hline
VpsKps\cite{tulsiani2015viewpoints} & 7 (F-RCNN\cite{ren2015faster}) & 0.713 & 0.647 \\
\hline
Ours & - & - & 46\\
\hline
\end{tabular}
\end{center}
\caption{Speed comparison.}
\label{tbl:speed}
\end{table}

\subsection{Household Dataset}\label{sec:philDataset}

\begin{table*}
\begin{center}
\begin{tabular}{|l||c|c|c|c|c||c|c|c|c|c|}
\hline
& \multicolumn{10}{|c|}{Scene 1}\\
\cline{2-11}
Method & \multicolumn{5}{|c||}{4 View AVP} & \multicolumn{5}{|c|}{8 View AVP}\\
\cline{2-11}
& chair & table & sofa & monitor & Avg & chair & table & sofa & monitor & Avg\\
\hline
Scratch & 11.9 & 15.2 & 35.4 & 49.4 & 28.0 & 16.3 & 9.9 & 31.8 & \textbf{45.1} & \textbf{25.8} \\
\hline
P3D & 2.9 & 17.4 & 37.4 & 18.3 & 19.0 & 2.1 & 11.1 & 25.2 & 6.1 & 11.2\\
\hline
P3D Finetuned & \textbf{13.4} & \textbf{26.4} & \textbf{43.9} & \textbf{53.0} & \textbf{34.2} & \textbf{20.0} & \textbf{12.1} & \textbf{33.7} & 24.6 & 22.6\\
\hline
Results on P3D+ & 10.8 & 38.5 & 41.8 & 55.8 & 36.7 & 9.1 & 34.2 & 37.2 & 46.7 & 31.8 \\
\hline\hline
& \multicolumn{10}{|c|}{Scene 2}\\
\cline{2-11}
 & \multicolumn{5}{|c||}{4 View AVP} & \multicolumn{5}{|c|}{8 View AVP}\\
\cline{2-11}
& chair & table & sofa & monitor & Avg & chair & table & sofa & monitor & Avg\\
\hline
Scratch & 18.1 & 36.7 & 44.5 & 45.3 & 36.1 & 13.1 & 14.8 & 42.1 & \textbf{38.1} & 27.0\\
\hline
P3D & 9.9 & 35.1 & 59.7 & 45.7 & 37.6 & 5.0 & 25.1 & 44.9 & 34.8 & 27.4\\
\hline
P3D Finetuned & \textbf{28.7} & \textbf{43.9} & \textbf{63.6} & \textbf{66.4} & \textbf{50.6} & \textbf{16.1} & \textbf{25.4} & \textbf{56.3} & 36.0 & \textbf{33.4} \\
\hline
Results on P3D+ & 10.8 & 38.5 & 41.8 & 55.8 & 36.7 & 9.1 & 34.2 & 37.2 & 46.7 & 31.8 \\
\hline\hline
& \multicolumn{10}{|c|}{Scene 3}\\
\cline{2-11}
 & \multicolumn{5}{|c||}{4 View AVP} & \multicolumn{5}{|c|}{8 View AVP}\\
\cline{2-11}
& chair & table & sofa & monitor & Avg & chair & table & sofa & monitor & Avg\\
\hline
Scratch & 1.8 & \textbf{28.8} & 46.3 & \textbf{66.1} & \textbf{35.7} & 0.6 & 9.9 & 32.7 & 25.8 & 17.3 \\
\hline
P3D & \textbf{7.6} & 13.0 & 41.4 & 47.7 & 27.4 & \textbf{8.0} & 6.2 & 31.1 & 30.5 & 18.9\\
\hline
P3D Finetuned & 5.3 & 27.7 & \textbf{48.9} & 56.3 & 34.6 & 4.4 & \textbf{11.1} & \textbf{36.3} & \textbf{38.0} & \textbf{22.5} \\
\hline
Results on P3D+ & 10.8 & 38.5 & 41.8 & 55.8 & 36.7 & 9.1 & 34.2 & 37.2 & 46.7 & 31.8 \\
\hline\hline
& \multicolumn{10}{|c|}{Scene 4}\\
\cline{2-11}
 & \multicolumn{5}{|c||}{4 View AVP} & \multicolumn{5}{|c|}{8 View AVP}\\
\cline{2-11}
& chair & table & sofa & monitor & Avg & chair & table & sofa & monitor & Avg\\
\hline
Scratch & 22.6 & 22.9 & - & 10.5 & 18.6 & 11.4 & 28.7 & - & 12.2 & 17.4 \\
\hline
P3D & \textbf{55.5} & \textbf{46.5} & - & 1.6 & \textbf{34.5} & \textbf{52.4} & \textbf{34.0} & - & 3.1 & 29.8\\
\hline
P3D Finetuned & 40.8 & 32.3 & - & \textbf{17.6} & 30.2 & 30.7 & 32.1 & - & \textbf{27.4} & \textbf{30.1} \\
\hline
Results on P3D+ & 10.8 & 38.5 & 41.8 & 55.8 & 36.7 & 9.1 & 34.2 & 37.2 & 46.7 & 31.8 \\
\hline\hline
& \multicolumn{10}{|c|}{Scene 5}\\
\cline{2-11}
 & \multicolumn{5}{|c||}{4 View AVP} & \multicolumn{5}{|c|}{8 View AVP}\\
\cline{2-11}
& chair & table & sofa & monitor & Avg & chair & table & sofa & monitor & Avg\\
\hline
Scratch & 21.8 & \textbf{21.2} & \textbf{47.0} & 21.4 & \textbf{27.9} & 11.0 & 5.7 & 34.4 & 7.3 & 14.6 \\
\hline
P3D & 6.2 & 18.3 & 41.2 & 21.0 & 21.7 & 4.7 & \textbf{9.0} & 27.3 & 14.4 & 13.8 \\
\hline
P3D Finetuned & \textbf{23.8} & 5.4 & 41.8 & \textbf{39.8} & 27.7 & \textbf{25.0} & 4.5 & \textbf{36.0} & \textbf{14.9} & \textbf{20.1} \\
\hline
Results on P3D+ & 10.8 & 38.5 & 41.8 & 55.8 & 36.7 & 9.1 & 34.2 & 37.2 & 46.7 & 31.8 \\
\hline
\end{tabular}
\end{center}
\caption{Results on the Household Dataset. Evaluation on the household data, results are shown for our model trained on (P3D), then fine-tuned on the household data (P3D+finetune), and for our model trained only on the household data (scratch).  We include the results of our P3D model on the P3D+ dataset in order to roughly compare the difficulty of the datasets for these categories.
}
\label{tbl:rohit}
\end{table*}

In addition to using the Pascal 3D+ dataset, we collected over 3,700 images using a Kinect v2 sensor in five real-world household scenes. These five scenes include a bedroom and four open kitchen/living rooms. From these scenes we label four of the Pascal categories: chair, dining table, monitor, and sofa. See the supplementary material for additional information regarding the number of images and objects in each scene. 

We are able to cheaply and easily collect labels for all images by first sparsely reconstructing each scene using COLMAP~\cite{schoenberger2016sfm}, and then densely reconstructing the scene using CMVS~\cite{furukawa2010cmvs,furukawa2007pmvs}. Following reconstruction, we label each object in the dense 3D point cloud and project the labeled point cloud to a bounding box in each image. Using depth data from the Kinect, we are able to automatically adjust the bounding boxes projected from the point cloud to account for occlusion. Moreover, we are also able to get consistent object pose information for every image in the scene by labeling just a single image per object and using the camera positions from the reconstruction to discern the pose of the object in every image.

\subsubsection{Experiments on Household Dataset}\label{sec:rohitexp}
%
Using our household dataset for real world scenes, we set out to evaluate how our detection and pose estimator trained on Pascal 3D+ performs on real-world scenes. Additionally, we compare training a detection and pose estimation model from scratch on the collected dataset (Scratch), using our detector trained on Pascal 3D+ (P3D), and finetuning the Pascal 3D+ detector on the training scenes (P3D Finetuned). When training our model on our Household Dataset, we train on four scenes and test on the held out scene.

Table \ref{tbl:rohit} presents the results of these experiments. Unsurprisingly, our model generally performs best when first trained on the Pascal 3D+ dataset then finetuned on labels from our real-world scenes. We provide qualitative results in the supplementary material.

\section{Conclusion}\label{sec:conclusion}
We have extended the fast SSD detector to estimate object pose. Our model achieves comparable accuracy to state-of-the-art methods on the standard Pascal 3D+ dataset\cite{xiang2014beyond}, while offering at least a 46x speedup. The ability to simultaneously detect and estimate the pose of an object opens the door for a variety of use cases. Additionally, we have collected a dataset consisting of five everyday scenes with object bounding boxes and poses labeled. Our results on this dataset demonstrate that the model proposed in this paper has the capacity to quickly and accurately perform simultaneous object detection and pose estimation in real-world scenes. 

\section*{Acknowledgements}
\noindent We would like to thank Jennifer Wu for helpful discussions and comments. This work was funded in part by a NSF Graduate Research Fellowship for PP, NSF NRI 1526367/1527208, and NSF 1452851.

{\small
\bibliographystyle{ieee}
\bibliography{egbib}
}

\end{document}